# Pixel-global Self-supervised Learning with Uncertainty-aware Context Stabilizer

Zhuangzhuang Zhang, Weixiong Zhang

**Introduction**

Most computer vision (CV) tasks can be classified into classification and dense prediction. Dense CV prediction tasks predict or estimate a label for every pixel of input images, such as depth prediction, semantic segmentation, and image registration [3, 4], each of which has a wide range of applications. The development of deep learning has brought significant improvements to the CV field; deep learning-based methods have been shown to outperform heuristic-based methods significantly. One prominent advantage of deep learning-based methods is that they learn to extract features rather than rely on hand-crafted features [5], unlocking the potential of learning salient features that benefit downstream tasks. While deep-learning methods can extract features from images automatically, they have a serious drawback of their enormous demand for training data. This high demand is rarely satisfied in the medical field because annotated datasets are expensive to prepare. The lack-of-annotated data is exacerbated for medical image applications because data annotation requires specialized medical expertise [6, 7].

A general hybrid learning approach has been proposed to address this critical annotation shortage, which combines self-supervised pre-training and supervised fine-tuning [8-11]. In the self-supervised pre-training phase, various tasks, namely pre-tasks, are introduced for a chosen backbone neural network to learn semantically meaningful representations. After pre-training, the backbone network is fine-tuned for downstream tasks. For example, a pre-task generated differently rotated images as training samples. The backbone network learns to predict the rotation applied, during which it learns to encode quality representations. After pre-training, the backbone network is fine-tuned for the downstream image classification task, in which the model learns to classify input images into different classes.

The latest SSL methods adopt contrastive representation learning [2]. Contrastive learning is based on the assumption that differently augmented views of the same image should have similar representation, and augmented views of different images should have distinct representations [10, 12]. A typical contrastive SSL method is structured as follows. Assuming we need to train a ResNet50 [13] (a classic deep learning backbone for images) for downstream image classification tasks, contrastive SSL sets up a student network and a teacher network with the same architecture (ResNet50). In each pre-training iteration, two differently augmented views of the same image are fed into the student and teacher network. We use the discrepancy between their output representations as the loss to update student network parameters. Teacher network parameters are normally updated by the moving average [14] of the student's parameters. While learning to perform well for this contrastive pre-task, the backbone network (ResNet50) learns to generate quality representations for input images. After pre-training the student's ability to extract features, we keep its parameters and fine-tune them with downstream classification datasets.

However, most contrastive SSL methods [3, 14-17] focus on image-level global consistency by treating every image as a class. However, local consistency between two augmented views of the same image is overlooked. Recent SSL models attempt to explore local consistency for generic images [18-20] and medical images [1, 2], yet they are still limited to enforcing the consistency at the region level. In these methods, each feature vector corresponds to a region in the original image. Enforcing such region-level consistency is too coarse-grained to be accurate or adequate for downstream pixel-wise prediction tasks, such as semantic segmentation and registration. While pixel-level consistency was explored earlier [1, 21], pixel-to-pixel consistency modeling has not been well studied. It has been studied by generative self-supervised learning methods [22] (comparison in Appendix A), but the potential of building pixel-level fine-grained SSL methods has not been explored. Furthermore, the context difference between a pixel in one view and its counterpart in the other view was not considered. For example, a pixel on the edge of one cropped view can be at the center of the other cropped view. Different data augmentations create this context gap, and it is not ideal to directly push the feature vector towards its corresponding one in the high-dimensional space.

We developed a novel SSL approach to capture global consistency and pixel-level local consistencies between differently augmented views of the same images to accommodate downstream discriminative and dense predictive tasks. We adopted the teacher-student architecture used in previous contrastive SSL methods [3, 14, 20]. In our method, the global consistency is enforced by aggregating the compressed representations of augmented views of the same image. The pixel-level consistency is enforced by pursuing similar representations for the same pixel in differently augmented views. Importantly, we introduced an uncertainty-aware context stabilizer to adaptively preserve the context gap created by the two views from different augmentations. Moreover, we used Monte Carlo dropout [23] in the stabilizer to measure uncertainty and adaptively balance the discrepancy between the representations of the same pixels in different views.

We experimentally tested and evaluated the new method using medical images, where high-quality annotated data are commonly unavailable or insufficient. For instance, a widely used general image dataset, ImageNet [4, 24], contains more than four million images. In contrast, the samples in most medical image datasets with annotation are in the orders of hundreds to thousands. None of the public medical image datasets (even with missing or incomplete annotations) has more than one million images. The data shortage is even more severe in cases where annotations are more labor-intensive, such as semantic segmentation. Our experiment assembled the largest dataset for medical image semantic segmentation, with 1808 cases. Despite the small samples size compared to generic image datasets such as ImageNet, we focused on image semantic segmentation as the downstream task to assess the performance of our new approach against several state-of-the-art methods. Thanks to its pixel-level consistency modeling and context stabilizer, the new approach showed superior performance over the existing methods.

In short, we make two main contributions in this paper:

- We propose a novel contrastive SSL approach that effectively enforces global and pixel-level consistencies, enabling deep learning models to automatically derive semantically meaningful representations and providing great transfer learning potential to various downstream tasks, e.g., semantic segmentation.

- We address the challenge of modeling pixel-level consistency by proposing an uncertainty-aware context stabilizer. This stabilizer has two eminent features. It adaptively cancels the context gap between two random data augmentations, which benefits pixel-wise consistency modeling and stabilizes the learning process by estimating uncertainty via Monte Carlo dropout.

**Related work**

Despite the success of deep learning techniques in computer vision, their demand for large quantities of training data and human supervision has been a bottleneck for applications where data annotations are limited or expensive [8]. To the rescue, SSL methods build models to learn pertinent image representations via completing different pre-tasks, guided by generated supervision signals.

SSL approaches can be summarized as in three categories based on the pre-task designs they adopt: predictive, generative, and contrastive SSL [12]. The predictive SSL methods apply classification pre-tasks in their models, training the backbone networks to make predictions based on learned latent features from unlabeled data. Early pre-tasks include image exemplar [25], relative position prediction [25], jigsaw puzzle [26], and rotation prediction [27]. The generative SSL methods use reconstruction pre-tasks for building models to learn latent features without human annotation. Generative pre-tasks include image denoising [28], inpainting [29], and colorization [30].

Recently, contrastive SSL has attracted great attention and led to various methods, such as contrastive predictive coding, simLR [3], MOCO [31], BYOL [14], and DINO [20]. The main idea of contrastive learning is to build positive pairs and negative pairs of examples via different ways of data augmentation and pair selection, and negative pairs may not always be necessary [14, 20]. Constructive SSL models learn to enforce positive pairs to be consistent while negative pairs to be dissimilar. A typical architecture of contrastive SSL has two branches: a student (online) branch and a teacher (target) branch, both of which use the same network backbone structures [1-3, 14, 18, 20, 31]. The outputs from the two branches are contrasted to assess the consistency between the two representations that the two branches learn. The student network parameters are updated by a back-propagation loss, whereas the teacher network parameters are normally updated by the moving average of the student network parameters [20, 31].

It is important to note that most existing contrastive SSL methods are limited to capturing global consistency between projected feature vectors that summarize the whole images [3, 14]. They only enforce the consistency between global feature vectors and do not capture the high-quality semantics of these images. To this end, dense contrastive learning methods have been proposed [18, 19] to enforce dense consistency between every pair of feature vectors on the extracted feature maps of the same images. However, they still focus on consistency between down-sampled feature maps [18, 19], enforcing region-to-region consistency (depending on how large each feature vector represents) rather than pixel-to-pixel consistency. For dense prediction tasks like semantic segmentation, modeling region-level consistency is too coarse-grained to be adequate.

In our new method, we enforced pixel-level consistency while maintaining global consistency to support classification and dense prediction tasks. The closest related works are [1, 21], but they have two major drawbacks. Firstly, they enforce local consistency without considering the context gap created by different data augmentations. This design is suboptimal because the representation for each pixel contains not just color and intensity information, but more importantly, its context.

The context information is even more crucial in single-channeled images like radiotherapy images. This context information is randomly shifted during different augmentations, and the context gap between branches is randomly created and overlooked in previous works [1, 18, 19, 21]. Our method overcomes this drawback by preserving the gap using our context stabilizer branch.

Secondly, directly enforcing consistency between each pixel representation makes the learning process unstable. Both student and teacher networks are randomly initialized, and the teacher network cannot create quality representations for all pixels in the early stages of learning. Hence, our method implements an uncertainty measuring module to stabilize the learning process.

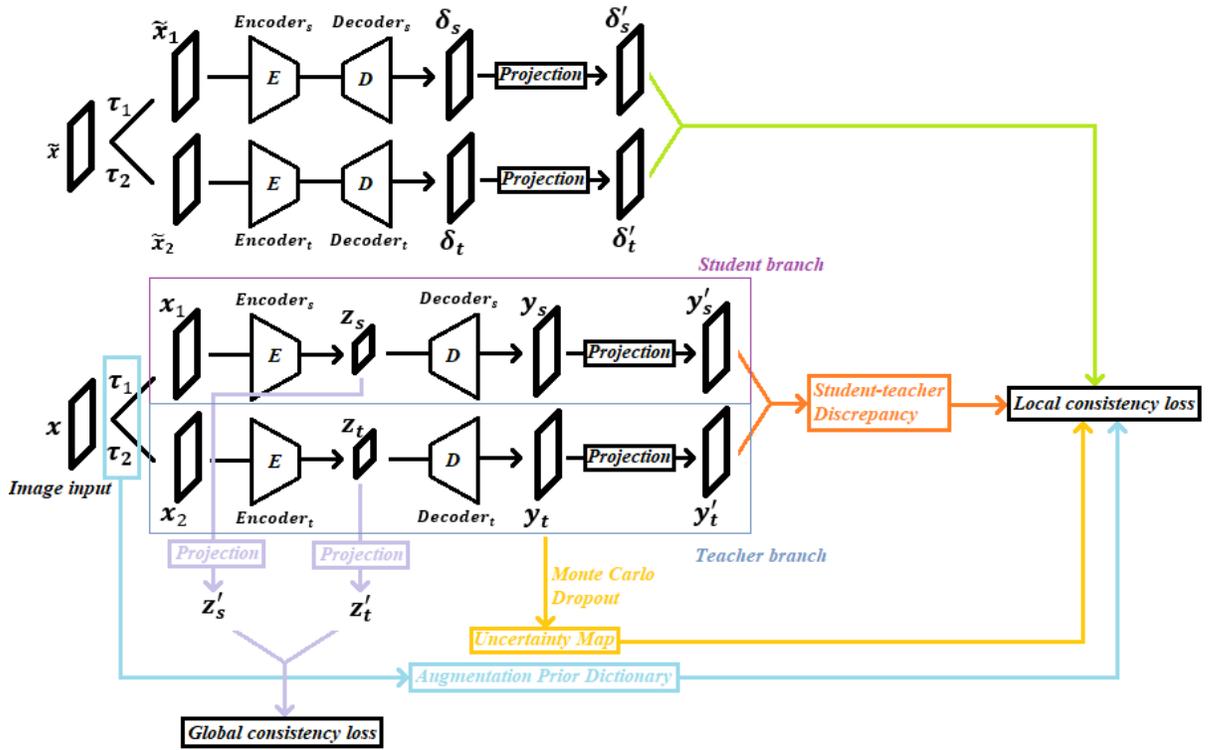

Figure 1: Overall structure of the proposed self-supervised learning framework. Input image $x$ goes through two random augmentations ($\tau_1, \tau_2$), then its augmented view $x_1$ and $x_2$ go through the student and teacher branch. $z_s$ and $z_t$ are features extracted by the encoder, and their projected feature vector are used to compute the global consistency loss. The output of the decoders are feature maps $y_s$ and $y_t$, and their projected views are $y_s'$ and $y_t'$, with which we compute the local consistency loss under the guidance of the augmentation prior dictionary, uncertainty map, and the context gap preservation. Details of the local consistency loss computation are in section 3.4.

**Methodology**

Our method adopts the student-teacher contrastive SSL architecture, in which the student and teacher networks share the same backbone. We capture the global consistency in the middle of the backbone via a projection layer and the pixel-wise local consistency between backbone output representations. We introduce a novel uncertainty-aware context stabilizer with two prominent features: 1) it uses training data to estimate the contextual gap created by different augmentations, and 2) it uses random dropout to measure uncertainty.

This section introduces the overall setup of the new approach and then discusses the global and pixel consistency modeling. The uncertainty-aware context stabilizer will be described along with the pixel-level consistency.

*Contrastive learning*

The proposed approach is illustrated in Figure 1. An input image $x$ goes through two random data augmentations $\tau_1$ and $\tau_2$ to produce two augmented views $x_1$ and $x_2$, respectively. Views $x_1$ and $x_2$ are then respectively feed to the student and teacher networks. Data augmentation operations include cropping & scaling, rotation, vertical & horizontal flipping, and contrast & gamma adjustment.

The student and teacher networks have the same architecture (Figure 1) and the initial weights of their parameters. Note that the new method can utilize any encode-decoder-style backbone [32]. The two encoders extract feature maps $z_s$ and $z_t$ from their inputs, which are then projected into feature vectors $z_s'$ and $z_t'$. We enforce the global consistency on $z_s'$ and $z_t'$ and the pixel-level local consistency on the projected network outputs $y_s'$ and $y_t'$. Therefor, the training loss for the student network is computed as:

$$\mathcal{L} = \ell_{global} + \ell_{local}. \tag{1}$$

Inspired by previous contrastive SSL methods [14, 20], we use the back-propagation loss to update the parameters $\theta_s$ of the student network. We employ an exponential moving average (EMA) to update the parameters $\theta_t$ of the teacher network:

$$\theta_t \leftarrow \lambda\theta_s + (1-\lambda)\theta_s, \tag{2}$$

where $\lambda$ follows a cosine schedule from 0.996 to 1 during training [14]. Mode collapse, in which student and teacher generate the same but meaningless representations, is avoided by the batch-normalization technique [33].

*Global consistency*

We achieve global consistency under the assumption that differently augmented views of the same image should have similar global representations [10, 14, 20]. We project the extracted feature maps $z_s$ and $z_t$ (output of the two encoders) to two global feature vectors $z_s{'}$ and $z_t{'}$. The projection modules summarize the features extracted by the encoders into two $1 \times 1$ feature vectors whose values are then aggregated by two softmax functions as suggested by [20]. We train the student network to predict the representation of the teacher network with a cross-entropy loss:

$$\ell_{global}(z_t', z_s') = -z_t' \log z_s'. \tag{3}$$

*Pixel-level consistency*

**Augmentation-guided pixel-to-pixel matching.** We enforce the pixel-level local consistency for corresponding pixels in the two augmented views [2, 19, 21]. Intuitively, for each pixel in the student's augmented view, the student learns to predict the teacher's representation of its corresponding pixel in the teacher's augmented view. This leads to a pixel-to-pixel registration problem across two augmented views. PGL [2] solves the problem by computing matching regions with augmentation priors, which works well with cropping, rotation. Our method keeps track of the augmentation priors with a dictionary to map corresponding pixel representations in $O(1)$ look-up time. The dictionary contains key-value pairs storing each pixel's location in augmented view 1 (key) and its location in augmented view 2 (value). Not all pixels in augmented view 1 have a matching pixel in augmented view 2 because of random cropping and scaling augmentation. An intuitive form of the local consistency loss would be:

$$\ell_{local}(y'_t, y'_s) = -\sum_i y'_{ti} \log y'_{si}, \tag{4}$$

in which $y'_t, y'_s$ are the output feature maps of the teacher and student networks, respectively; $y'_{ti}, y'_{si}$ are the feature vectors of pixel $i$ on the output feature maps of the teacher and student networks.

We only compute the discrepancy between the two pixels' representations because matching pixels across differently augmented views are not guaranteed due to random cropping and scaling.

**Context gap stabilizer.** We emphasize that the pixel representation encodes not only the color and intensity of every pixel but also the context information of the pixel, such as its position in the image and its relative position to other pixels. This context gap is created by two random data augmentations, usually not uniform among all pixels considering their different original locations. It is randomly created and latent, so we propose to cancel this gap from our loss using another training image. We randomly sample another training image $\tilde{x}$ and apply the same data augmentations. We then compute the discrepancy between corresponding locations from the representation maps of $\tilde{x}$. The pixel-to-pixel registration problem across two augmented views can be solved by the same prior dictionary we mentioned earlier. We refine the local consistency modeling with this term, emphasizing the significance of pixel context information. This context stabilizer also smoothens the loss curve by adaptively negating the gap created by two random data augmentations. We revise the local consistency as:

$$\ell_{local}(y'_t, y'_s) = |\sum_i -y'_{ti} \log y'_{si} + \delta'_{ti} \log \delta'_{si}|, \tag{5}$$

in which $\delta'_{ti}$ and $\delta'_{si}$ are projected representations from the teacher and student networks.

**Adaptive uncertainty weighting.** In our contrastive learning scheme, the student learns to predict the representation generated by the teacher network. However, the teacher network is initialized randomly, which means it cannot produce trustworthy representations in the early stages of learning. We address this problem by measuring the uncertainty of the teacher network and letting the student network learn from the teacher based on how certain the teacher is. We implement the Monte Carlo Dropout technique [34] to compute a map of the uncertainty of the teacher $\boldsymbol{U_t}$, with which we weight the pixel representation discrepancy for each location. We perform multiple forward passes for the teacher network with dropout layers activated and compute the average prediction entropy for every location. Intuitively, for each location, if the teacher's projected output is close to 0 or 1, meaning the entropy is close to 0, it indicates that the teacher is more certain about its output, then the student will learn more from its error by weighting it more highly than less certain locations. The idea was adopted by supervised learning methods [23]. Like a human, the model usually "feels" uncertain on the boundary area (Figure 2) due to low-contrast and fuzzy boundaries, and a previous method [23] discards the loss for regions where the model's certainty is under a chosen threshold. We propose to switch from hard filtering to soft weighting, eliminating the work for tuning this threshold. We choose to perform 20 forward passes on the teacher branch for each training iteration considering the trade-off between the uncertainty estimation quality and computation cost. The complete local consistency loss of our method is:

$$\ell_{local}(\boldsymbol{y'_t}, \boldsymbol{y'_s}) = \sum_i (1 - \boldsymbol{U_{ti}})(|-\boldsymbol{y'_{ti}} \log \boldsymbol{y'_{si}} + \boldsymbol{\delta'_{ti}} \log \boldsymbol{\delta'_{si}}|), \qquad (6)$$

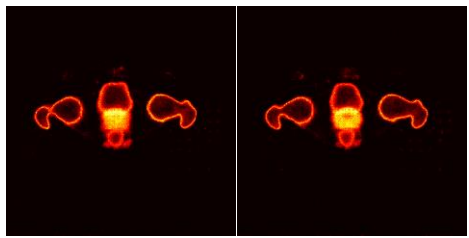

Figure 2: Example uncertainty maps of the model's prediction on pelvic CT images. Higher intensity indicates higher uncertainty.

in which $U_{ti}$ is the teacher's uncertainty at location $i$.

**Experimental results**

*Datasets*

**Pre-training datasets.** We collected 1808 CT scans to pre-train our method and two state-of-the-art methods [1, 2]. This ensembled dataset with roughly 0.4 million images is the largest pre-training dataset for medical images to our knowledge. Nonetheless, it is still only 1/10 of ImageNet [24] (more than four million images). The pre-training dataset is collected from five public sources:

- 660 lung CT scans from the RibFrac dataset [35].

- 1148 CT scans from subsets of the Medical Segmentation Decathlon (MSD) challenge [36], including Hepatic Vessel, Colon Tumor, Pancreas, and Lung Tumor.

**Datasets for downstream tasks.** Our proposed framework enforces image-level global consistency and pixel-level local consistency, emphasizing its prominent ability for dense context modeling. Hence we collected three CT segmentation dataset across different sites to evaluate the pre-training performance of the framework:

- **KiTs Dataset** [37]: The dataset has 210 abdomen CT scans with segmentation for kidney and tumor. Ground-truth contours are drawn by medical students supervised by their clinical chair [2]. We split the dataset as 80/25/105 cases (6143/2396/7794 images) for training/validation/testing.

- **Pelvic CT Dataset** [38]: This in-house dataset has 100 pelvic CT scans with segmentation for prostate (PRT), bladder (BLD), rectum (RCT), left femur (LF), and right femur (RF). Ground truths are the consensus contours drawn by two radiation oncologists with over ten years of experience [7]. We split the dataset as 40/10/50 cases (2560/640/3200 images) for training/validation/testing.

- **Head-and-neck CTs Dataset** [39, 40]**:** This is another in-house dataset having 100 head-and-neck CT scans with segmentation for brain stem (BS), chiasm (CM), mandible (MD), left/right optic nerve (LO/RO), left/right submandibular (LS/RS). The left/right parotid (LP/RP) Ground truths for the dataset are the consensus contours drawn by two radiation oncologists with over ten years of experience. We split the dataset as 40/10/50 cases (2560/640/3200 images) for training/validation/testing.

The three downstream datasets are large ones of the same applications, yet they are still tiny compared to general image datasets. Our work harnesses the power of self-supervised pre-training to boost the learning outcomes on small downstream datasets significantly.

*Implementation*

**SSL pre-training**. We used the ResUnet50 [7, 13] backbone for experiments. For the pre-training dataset (1808 CT scans), we clipped the pixel values to the range of [-1024, +325] of Hounsfield units (HU). We then used the Z-score normalization and center-cropped each slice to the size of $384 \times 384$, considering the computational and spatial complexity. We randomly cropped and scaled every input image to $224 \times 224$ as the first augmentation for both student and teacher networks. Adopting the training configuration of BYOL [14], we use a LARS optimizer with a cosine decaying learning rate and warm-up period of 2 epochs. We set the initial learning rate to 0.2, batch size to 8, and the maximum epoch to 500 [2]. The decay rate of the teacher network was also set to increase from 0.996 to 1 [2, 14].

**Fine-tuning.** The same techniques used for the pre-training step were adopted for downstream training dataset augmentation. The input size was set to $384 \times 384$ for all three downstream datasets. We used the Adam optimizer [41] with a polynomial learning scheduler [42]. We set the initial learning rate to 0.001, batch size to 8, and maximum epoch to 100. We also applied an early stopping mechanism [43] to avoid overfitting. We evaluated the segmentation results with two widely used metrics for medical image segmentation: the Dice similarity coefficient (DSC) and Hausdorff Distance (HD) [44]. DSC measures the overlapping ratio of the prediction masks and the ground-truth masks; the larger the value is, the better. HD measures the distance between predicted contours and the ground-truth contours; the smaller the value is, the better.

*Comparison to the state-of-the-art methods*

We compared our proposed approach to three state-of-the-art SSL methods for medical images:

- Global and local SSL (GL) [1]: A SimCLR [3] variation that captures both global and local consistency, in which the local consistency is limited to divided regions of the decoder output feature maps.

- Prior-guided Local SSL (PGL) [2]: A BYOL [14] variation that emphasizes local consistency with augmentation prior guidance.

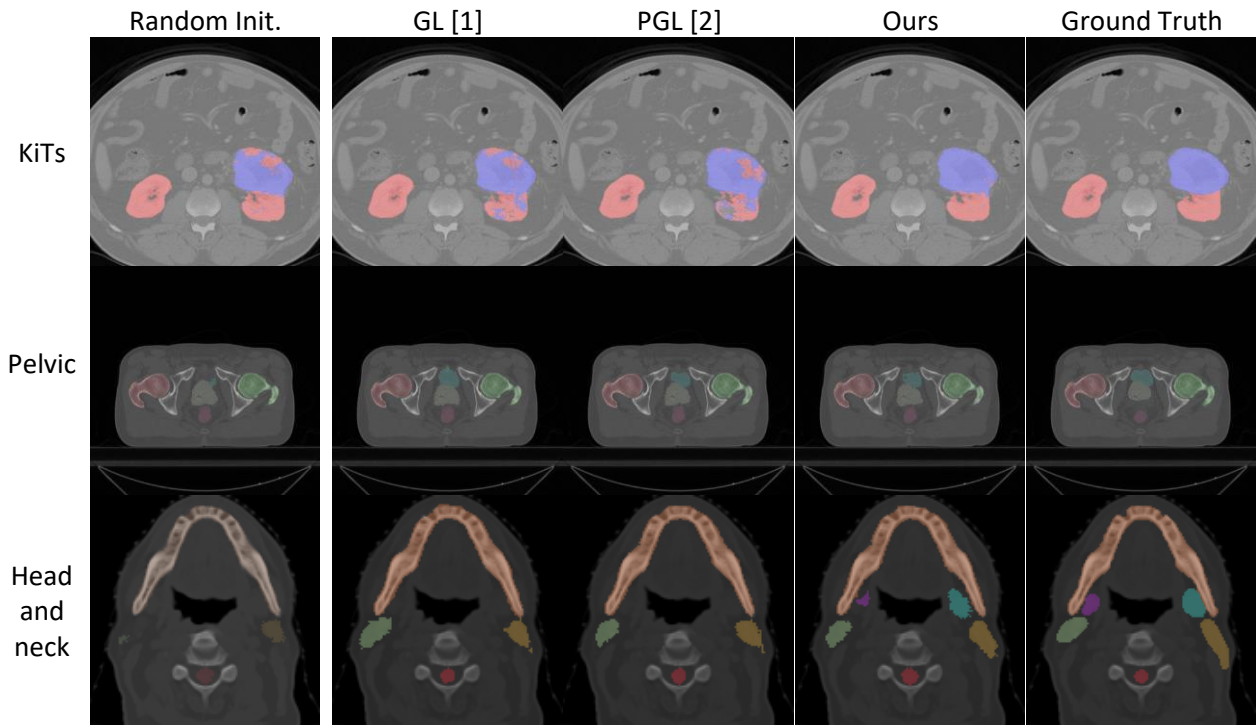

Figure 3: Visual comparison of three methods against the ground truth for medical image segmentation.

Table 1.a: Results comparison on the KiTs dataset (7794 samples). Statistical significance scores are calculated based on DSCs, and significant improvements (*p-value*<0.05) over compared pre-training methods are marked with '*'.

| Pre-train | Metrics | Organ | Tumor | Avg. |
|---|---|---|---|---|
| Random Initialization | DSC | 90.30 | 53.14 | 71.72 |
| | HD (mm) | 30.94 | 68.51 | 49.73 |
| | *p-value* | | <0.01* | |
| Global and local [1] | DSC | 91.52 | 61.74 | 76.63 |
| | HD (mm) | 25.72 | 73.51 | 49.62 |
| | *p-value* | | <0.01* | |
| PGL [2] | DSC | **93.72** | 63.65 | 78.69 |
| | HD (mm) | **23.33** | 69.53 | 46.43 |
| | *p-value* | | <0.01* | |
| Ours | DSC | 92.29 | **72.57** | **82.43** |
| | HD (mm) | 25.29 | **65.58** | **45.44** |

Comparison results on three downstream datasets are shown in Table 1. We compared the Dice similarity coefficients (DSC) and Hausdorff distances (HD) of the three methods for each class and the average scores of all classes, except the background class. Compared to randomly initialized ResUnet50, our proposed SSL pre-training achieved a DSC increase of 14.93%, 4.19%, 12.52%, and an HD decrease of 8.63%, 21.03%, 32.19% on the KiTs, pelvic, and head-and-neck datasets, respectively. Such improvements indicate significantly better segmentation quality in radiotherapy treatment, hence better dose prediction and treatment outcome. It also beats GL and PGL by significant margins, achieving state-of-the-art performance. A brief visual comparison is in Figure 3.

Table 1.b: Results comparison on the in-house pelvic dataset (3200 samples).

| Pre-train | Metrics | PRT | BLD | RCT | LF | RF | Avg. |
|---|---|---|---|---|---|---|---|
| Random Initialization | DSC | 66.90 | 93.25 | 76.33 | 95.87 | 96.32 | 85.73 |
| | HD (mm) | 9.28 | 9.37 | 7.32 | 5.98 | 5.87 | 7.56 |
| | p-value | | | <0.01* | | | |
| Global and local [1] | DSC | 68.54 | 94.01 | 77.11 | **96.34** | 96.12 | 86.42 |
| | HD (mm) | 8.01 | 8.56 | 8.50 | 4.76 | 4.52 | 6.87 |
| | p-value | | | <0.01* | | | |
| PGL [2] | DSC | 69.00 | **95.16** | 78.25 | 96.33 | **96.51** | 87.05 |
| | HD (mm) | 7.44 | 8.49 | **5.49** | 4.36 | **4.37** | 6.03 |
| | p-value | <0.01* | 0.55 | | <0.01* | | |
| Ours | DSC | **79.14** | 95.01 | **81.63** | 95.36 | 95.48 | **89.32** |
| | HD (mm) | **7.23** | **8.02** | 5.50 | **4.19** | 4.89 | **5.97** |

Table 1.c: Results comparison on the in-house head-and-neck dataset (3200 samples).

| Pre-train | Metrics | BS | CM | MD | LO | RO | LS | RS | LP | RP | Avg. |
|---|---|---|---|---|---|---|---|---|---|---|---|
| Random Initialization | DSC | 80.22 | 5.54 | 88.12 | **32.92** | 23.86 | 64.09 | 64.01 | 78.29 | 66.36 | 55.93* |
| | HD (mm) | 3.86 | 9.98 | 7.70 | 13.08 | 15.57 | 8.24 | 6.67 | 8.07 | 8.42 | 9.07 |
| | p-value | | | | | <0.01* | | | | | |
| Global and local [1] | DSC | 83.22 | 3.23 | 89.71 | 22.0 | 15.01 | 67.13 | 68.17 | 76.46 | 76.53 | 55.72* |
| | HD (mm) | 3.65 | 10.30 | 6.22 | 7.86 | 8.52 | **6.21** | 8.24 | 7.55 | 7.13 | 7.30 |
| | p-value | | | | | <0.01* | | | | | |
| PGL [2] | DSC | 83.82 | 3.24 | 89.89 | 24.13 | 19.27 | 59.36 | 70.24 | 77.50 | 77.89 | 56.15* |
| | HD (mm) | 3.59 | 9.39 | 5.50 | 7.39 | **8.18** | 6.25 | 8.62 | 6.94 | 6.83 | 6.97 |
| | p-value | | | | | <0.01* | | | | | |
| Ours | DSC | **86.80** | **9.99** | **91.12** | 29.50 | **27.52** | **77.11** | **77.08** | **83.26** | **84.00** | **62.93** |
| | HD (mm) | **2.98** | **7.80** | **4.13** | **7.29** | 9.21 | 7.99 | **4.65** | **5.93** | **5.35** | **6.15** |

*Pixel-wise consistency modeling evaluation*

We further studied the pixel-wise consistency modeling of our proposed SSL approach by breaking it down into three components: the augmentation-guided pixel-to-pixel matching, context gap stabilizer, and adaptive uncertainty weighting. The learning scheme with global consistency alone was used as the baseline for this analysis, similar to DINO [20]. Performance gains were observed on all three downstream datasets after adding each ingredient of the new pixel-wise consistency modeling (Figure 4). Moreover, the effect of each ingredient differed for different datasets; the more difficult the task is (e.g., various shapes and sizes of the organ, low-contrast, fuzzy, and ambiguous boundaries), the more gains we expected to have using the new pixel-to-pixel consistency modeling.

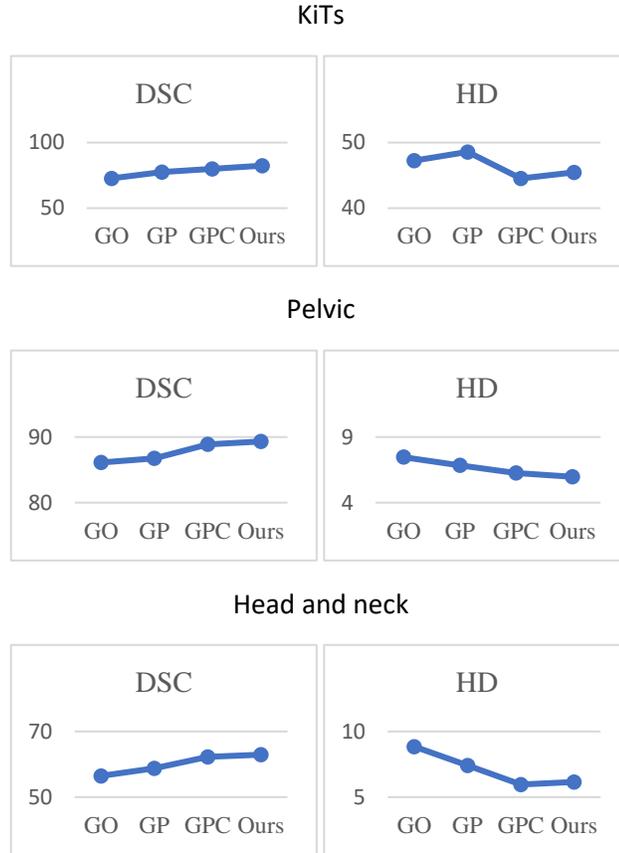

Figure 4: Ablation study on the components of our pixel-wise consistency modeling. "GO" is "global only"; "GP" is "GO" plus pixel-to-pixel matching"; "GPC" is "GP" plus the context gap stabilizer; "Ours" is "GPC" plus the adaptive uncertainty weighting.

## Discussion

### *Semi-supervised mode*

We evaluated the contrastive learning performance without extra pre-training datasets. With an annotated training dataset, we first run the contrastive self-supervised learning mode for model weight initialization, then start supervised training from these pre-trained weights. This mode learns from both unannotated data and annotated data, hence semi-supervised. Results (Table 2) showed that the semi-supervised mode of our method increases the DSC by 5.74%, 1.34%, and 7.24%;

Table 2.a: Results comparison on the KiTs dataset. Statistically significant differences (p-value<0.05) are marked with '*'. The number of samples is 7794.

| Pre-train | Metrics | Organ | Tumor | Avg. |
|---|---|---|---|---|
| Random Initialization | DSC | 90.30 | 53.14 | 71.72 |
| | HD (mm) | 30.94 | 68.51 | 49.73 |
| Ours | DSC | 92.36 | 59.32 | 75.84* |
| | HD (mm) | 23.52 | 66.61 | 45.07 |

Table 2.b: Results comparison on the in-house pelvic dataset. The number of samples is 3200.

| Pre-train | Metrics | Prostate | Bladder | Rectum | Left femur | Right femur | Avg. |
|---|---|---|---|---|---|---|---|
| Random Initialization | DSC | 66.90 | 93.25 | 76.33 | 95.87 | 96.32 | 85.73 |
| | HD (mm) | 9.28 | 9.37 | 7.32 | 5.98 | 5.87 | 7.56 |
| Ours | DSC | 70.32 | 93.21 | 78.01 | 96.32 | 96.52 | 86.88* |
| | HD (mm) | 8.31 | 8.12 | 6.69 | 5.64 | 5.82 | 6.92 |

Table 2.c: Results comparison on the in-house head-and-neck dataset. The number of samples is 3200.

| Pre-train | Metrics | BS | CM | MD | LO | RO | LS | RS | LP | RP | Avg. |
|---|---|---|---|---|---|---|---|---|---|---|---|
| Random Initialization | DSC | 80.22 | 5.54 | 88.12 | 32.92 | 23.86 | 64.09 | 64.01 | 78.29 | 66.36 | 55.93 |
| | HD (mm) | 3.86 | 9.98 | 7.70 | 13.08 | 15.57 | 8.24 | 6.67 | 8.07 | 8.42 | 9.07 |
| Ours | DSC | 82.02 | 6.05 | 88.29 | 33.33 | 25.24 | 69.19 | 69.32 | 79.15 | 78.25 | 58.98* |
| | HD (mm) | 4.01 | 9.32 | 7.21 | 9.68 | 10.78 | 7.51 | 6.66 | 7.28 | 7.59 | 7.78 |

decreases HD by 9.37%, 8.47%, and 14.22% for KiTs, pelvic, and head-and-neck datasets, respectively. This indicates that our proposed method boosts performance when no extra pre-training data is available. Supervised learning has been known to learn weaker representation than self-supervised models because supervision offers a shortcut [20]. Models learn to "guess" the right answers rather than understand the subject matter. Our self-supervised learning step prepares the model with great feature extraction ability, providing a great starting point for the supervised model.

*Robustness in low-resource settings*

We studied the robustness of our method in a low-resource setting by testing the model's performance while partially holding out the training data for fine-tuning (Figure 5). The results showed that our pre-trained model maintained 82.39%, 89.85%, and 80.25% of the DSC score with only 10% annotated training data for KiTs, pelvic, and head-and-neck datasets, respectively. This is significant because less than 10 cases (8 cases for KiTs, 4 cases for pelvic dataset, and head-and-neck dataset) were used to fine-tune the downstream segmentation tasks, affordable in most applications. Our method substantially relieves the thirst for annotated data for a wide range of deep learning-based medical imaging applications.

**Conclusion**

We developed a novel self-supervised learning method to train backbone networks to learn high-quality image representations that can be exploited for many downstream image processing tasks. The method enforces global and local consistencies between the student and teacher networks of the contrastive learning architecture. More crucially, it introduces an uncertainty-aware context gap estimator in pixel-to-pixel consistency modeling. The implemented new method is tailored to explicitly prepare the backbone for dense prediction types of downstream tasks, like depth prediction and semantic segmentation.

We evaluated the proposed method by conducting extensive experiments on three medical image segmentation datasets, covering 15 organs and tumors. The pre-training and downstream datasets we used are the largest for the applications to our knowledge. However, they are still tiny compared to general images, indicating the severe data shortage in medical image applications.

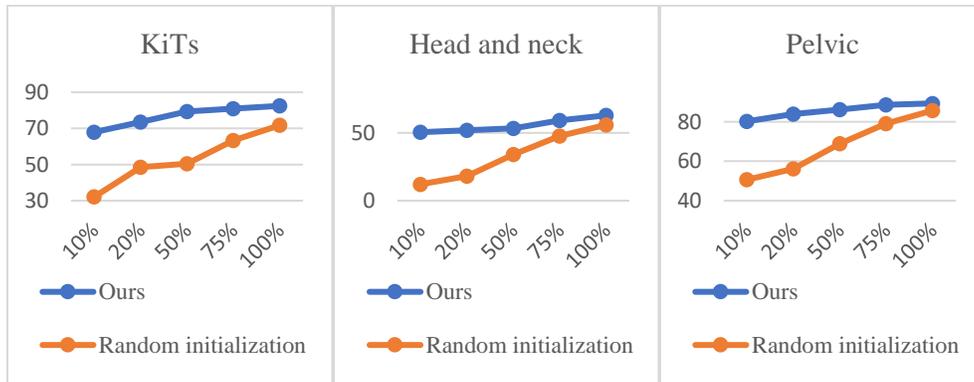

Figure 5: Robustness of our pre-trained model when fine-tune training dataset is partially held out.

We compared our method with two state-of-the-art methods. The results showed that our pixel-global SSL method achieved outperformed GL [1] and PGL [2] with significant margins. An ablation study has also been performed to prove the effectiveness of our context gap estimator and the adaptive uncertainty weighing module. The method can also be extended to semi-supervised and boost the model performance without additional pre-training datasets. While the current study focused on medical images, the new method is readily applicable to general images. We will further test its potential on general images in the future.

**Reference**

1. Chaitanya, K., et al., *Contrastive learning of global and local features for medical image segmentation with limited annotations.* arXiv preprint arXiv:2006.10511, 2020.
2. Xie, Y., et al., *Pgl: Prior-guided local self-supervised learning for 3d medical image segmentation.* arXiv preprint arXiv:2011.12640, 2020.
3. Chen, T., et al. *A simple framework for contrastive learning of visual representations.* in *International conference on machine learning.* 2020. PMLR.
4. Altaf, F., et al., *Going deep in medical image analysis: concepts, methods, challenges, and future directions.* IEEE Access, 2019. **7**: p. 99540-99572.
5. Ker, J., et al., *Deep learning applications in medical image analysis.* Ieee Access, 2017. **6**: p. 9375-9389.
6. Wu, Y., et al., *Semi-supervised Left Atrium Segmentation with Mutual Consistency Training.* arXiv preprint arXiv:2103.02911, 2021.
7. Zhang, Z., et al., *Semi-supervised semantic segmentation of prostate and organs-at-risk on 3D pelvic CT images.* Biomedical Physics & Engineering Express, 2021. **7**.
8. Dhere, A. and J. Sivaswamy, *Self-Supervised Learning for Segmentation.* arXiv preprint arXiv:2101.05456, 2021.
9. Chen, L., et al., *Self-supervised learning for medical image analysis using image context restoration.* Medical image analysis, 2019. **58**: p. 101539.
10. Jing, L. and Y. Tian, *Self-supervised visual feature learning with deep neural networks: A survey.* IEEE transactions on pattern analysis and machine intelligence, 2020.
11. Zhang, Z., B. Sun, and W. Zhang, *Pyramid medical transformer for medical image segmentation.* arXiv preprint arXiv:2104.14702, 2021.
12. Shurrab, S. and R. Duwairi, *Self-supervised learning methods and applications in medical imaging analysis: A survey.* arXiv preprint arXiv:2109.08685, 2021.
13. He, K., et al. *Deep residual learning for image recognition.* in *Proceedings of the IEEE conference on computer vision and pattern recognition.* 2016.
14. Grill, J.-B., et al., *Bootstrap your own latent: A new approach to self-supervised learning.* arXiv preprint arXiv:2006.07733, 2020.
15. Liu, C., et al., *Multiview Self-Supervised Segmentation for OARs Delineation in Radiotherapy.* Evidence-Based Complementary and Alternative Medicine, 2021. **2021**.
16. Xie, X., et al. *Instance-aware self-supervised learning for nuclei segmentation.* in *International Conference on Medical Image Computing and Computer-Assisted Intervention.* 2020. Springer.

17. Li, J., T. Lin, and Y. Xu. *SSLP: Spatial Guided Self-supervised Learning on Pathological Images*. in *International Conference on Medical Image Computing and Computer-Assisted Intervention*. 2021. Springer.
18. Wang, X., et al. *Dense contrastive learning for self-supervised visual pre-training*. in *Proceedings of the IEEE/CVF Conference on Computer Vision and Pattern Recognition*. 2021.
19. Xie, Z., et al. *Propagate yourself: Exploring pixel-level consistency for unsupervised visual representation learning*. in *Proceedings of the IEEE/CVF Conference on Computer Vision and Pattern Recognition*. 2021.
20. Caron, M., et al., *Emerging properties in self-supervised vision transformers.* arXiv preprint arXiv:2104.14294, 2021.
21. Yan, K., et al., *Self-supervised learning of pixel-wise anatomical embeddings in radiological images.* arXiv preprint arXiv:2012.02383, 2020.
22. Xia, Z. and A. Chakrabarti, *Training image estimators without image ground-truth.* arXiv preprint arXiv:1906.05775, 2019.
23. Yu, L., et al. *Uncertainty-aware self-ensembling model for semi-supervised 3D left atrium segmentation*. in *International Conference on Medical Image Computing and Computer-Assisted Intervention*. 2019. Springer.
24. Krizhevsky, A., I. Sutskever, and G.E. Hinton, *Imagenet classification with deep convolutional neural networks.* Advances in neural information processing systems, 2012. **25**: p. 1097-1105.
25. Dosovitskiy, A., et al., *Discriminative unsupervised feature learning with exemplar convolutional neural networks.* IEEE transactions on pattern analysis and machine intelligence, 2015. **38**(9): p. 1734-1747.
26. Noroozi, M. and P. Favaro. *Unsupervised learning of visual representations by solving jigsaw puzzles*. in *European conference on computer vision*. 2016. Springer.
27. Gidaris, S., P. Singh, and N. Komodakis, *Unsupervised representation learning by predicting image rotations.* arXiv preprint arXiv:1803.07728, 2018.
28. Vincent, P., et al. *Extracting and composing robust features with denoising autoencoders*. in *Proceedings of the 25th international conference on Machine learning*. 2008.
29. Pathak, D., et al. *Context encoders: Feature learning by inpainting*. in *Proceedings of the IEEE conference on computer vision and pattern recognition*. 2016.
30. Zhang, R., P. Isola, and A.A. Efros. *Colorful image colorization*. in *European conference on computer vision*. 2016. Springer.
31. He, K., et al. *Momentum contrast for unsupervised visual representation learning*. in *Proceedings of the IEEE/CVF Conference on Computer Vision and Pattern Recognition*. 2020.
32. Ronneberger, O., P. Fischer, and T. Brox. *U-net: Convolutional networks for biomedical image segmentation*. in *International Conference on Medical image computing and computer-assisted intervention*. 2015. Springer.
33. Richemond, P.H., et al., *BYOL works even without batch statistics.* arXiv preprint arXiv:2010.10241, 2020.
34. Kendall, A. and Y. Gal, *What uncertainties do we need in bayesian deep learning for computer vision?* arXiv preprint arXiv:1703.04977, 2017.
35. Jin, L., et al., *Deep-learning-assisted detection and segmentation of rib fractures from CT scans: Development and validation of FracNet.* EBioMedicine, 2020. **62**: p. 103106.
36. Simpson, A.L., et al., *A large annotated medical image dataset for the development and evaluation of segmentation algorithms.* arXiv preprint arXiv:1902.09063, 2019.
37. Heller, N., et al., *The kits19 challenge data: 300 kidney tumor cases with clinical context, ct semantic segmentations, and surgical outcomes.* arXiv preprint arXiv:1904.00445, 2019.

38. Zhang, Z., et al., *ARPM-net: A novel CNN-based adversarial method with Markov random field enhancement for prostate and organs at risk segmentation in pelvic CT images.* Medical physics, 2021. **48**(1): p. 227-237.
39. Zhang, Z., et al., *Weaving attention U-net: A novel hybrid CNN and attention-based method for organs-at-risk segmentation in head and neck CT images.* Medical Physics, 2021.
40. Zhang, Z., et al., *Semi-supervised semantic segmentation of prostate and organs-at-risk on 3D pelvic CT images.* Biomedical Physics & Engineering Express, 2021. **7**(6): p. 065023 %@ 2057-1976.
41. Zhang, Z. *Improved adam optimizer for deep neural networks.* in *2018 IEEE/ACM 26th International Symposium on Quality of Service (IWQoS).* 2018. IEEE.
42. Mishra, P. and K. Sarawadekar. *Polynomial learning rate policy with warm restart for deep neural network.* in *TENCON 2019-2019 IEEE Region 10 Conference (TENCON).* 2019. IEEE.
43. Prechelt, L., *Early stopping-but when?*, in *Neural Networks: Tricks of the trade.* 1998, Springer. p. 55-69.
44. Thada, V. and V. Jaglan, *Comparison of jaccard, dice, cosine similarity coefficient to find best fitness value for web retrieved documents using genetic algorithm.* International Journal of Innovations in Engineering and Technology, 2013. **2**(4): p. 202-205.